\documentclass[sigconf]{acmart} 

\AtBeginDocument{%
  \providecommand\BibTeX{{%
    \normalfont B\kern-0.5em{\scshape i\kern-0.25em b}\kern-0.8em\TeX}}}
\setcopyright{acmcopyright}
\copyrightyear{2018}
\acmYear{2018}
\acmDOI{XXXXXXX.XXXXXXX}

\acmConference[Conference acronym 'XX]{Make sure to enter the correct
  conference title from your rights confirmation emai}{June 03--05,
  2018}{Woodstock, NY}
  
\setcopyright{acmcopyright}
\copyrightyear{2023}
\acmYear{2023}
\acmDOI{XXXXXXX.XXXXXXX}

\acmConference[CONCATENATE - HRI]{}{March 13, 2023}{Stockholm, SE}

\acmPrice{15.00}
\acmISBN{978-1-4503-XXXX-X/18/06}

\begin{document}

\title{One Size Doesn't Fit All:\\ Personalised Affordance Design for Social Robots}

\author{Guanyu Huang}
\affiliation{%
  \institution{University of Sheffield}
  \city{Sheffield}
  \country{United Kingdom}
}
\email{ghuang10@sheffield.ac.uk}
\orcid{0000-0002-0266-6987}

\author{Roger K. Moore}
\affiliation{%
  \institution{University of Sheffield}
  \city{Sheffield}
  \country{United Kingdom}
}
\email{r.k.moore@sheffield.ac.uk}
\orcid{0000-0003-0065-3311}


\begin{abstract}
In human-robot interaction, personalisation is essential to achieve more acceptable and effective results. Placing users in the central role, many studies have focused on enhancing social robots’ abilities to perceive and understand users. However, little is known about improving users’ perceptions and interpretation of a social robot in spoken interactions. The work described in the paper aims to find out what affects the personalisation of a social robot’s affordance, namely appearance, voice and language behaviours. The experimental data presented here is based on an ongoing project. It demonstrates the many and varied ways in which people’s preferences for a social robot’s affordance differ under different circumstances. It also examines the relationship between such preferences and people’s expectations of a social robot’s characteristics like competence and warmth. It also shows that individuals have different perceptions of the same robot’s language behaviours. These results demonstrate that one-sized personalisation does not fit all. Personalisation should be considered a comprehensive approach, including appropriate affordance design, to suit the users’ expectations of social roles.
\end{abstract}




\maketitle

\section{Introduction}
Personalisation is a type of adaptation. Personalised human-robot interaction aims to improve the effectiveness and acceptability of a robot by adapting a robot’s behaviours to interact with a specific individual or group. Such adjustment is based on physical and social factors in the given environment where the interaction takes place, as well as users’ preferences, behaviours, and characteristics \cite{hellou2021personalization}. Along with the development of robots in social domains, research for personalised HRI has focused on making robots learn and adapt to users. These adaptations can be a long-term process \cite{portugal2015socialrobot, gockley2005designing, lee2012personalization} and sometimes culturally sensitive \cite{tuyen2018emotionalindividuallong, kanda2010communication}. For short-term individual interaction, most studies have paid attention to enhancing robots’ abilities to perceive and interpret users’ attention \cite{mccoll2013meal}, personality \cite{aly2013model}, behaviour \cite{gross2009toomas} and abilities \cite{torrey2006effects}. 

Whilst it is important to develop a robot's abilities to adapt its behaviours, it is also worth paying attention to a social robot's static features which do not change over the duration of interactions, such as its appearance and voice. Such features help form the first impression, a fundamental factor in human-human encounters with regard to people's social cognition (e.g., trust and rapport) \cite{Ambady2008firstimpression}. Also, they are related to a robot's affordance, namely its perceptual action possibilities. According to the Affordance Theory\cite{gibson1977theory,matei2020affordance}, affordance shapes people's perception of an object and affects people’s behaviours. Thus, it is important to design appropriate affordances for robots to develop a satisfactory first impression and optimise users' experiences in the HRI.

Two dimensions widely used to measure people's social perceptions are warmth and competence \cite{cuddy2008warmth,he2019stereotypes}. The former captures the perceived friendliness and good intention. The latter captures the perceived ability to deliver on those intentions. Given that social robots are used in social domains to play social roles, it makes sense to investigate what people expect robots to be like in different roles. A well-developed model to measure warmth and competence in social roles is the stereotype content model (SCM), which has been proven reliable in many experimental tests across different cultural contexts \cite{cuddy2008warmth,russell2008s,cuddy2009stereotype}. It has also been used in studies of how people perceive virtual agents (e.g., \cite{bergmann2012second}).

This paper adapts this model within a programme of empirical experimental work. The experiment firstly aims to find out what people's expectations of social robots' warmth and competence are in given social scenarios and how such expectations affect people's preferences for social robots' affordances. Secondly, it aims to identify personalised affordance design principles by studying people’s preferences and perceptions of a social robot’s affordance. 

\section{Research Hypothesis and Questions}
The main hypothesis is that a social robot’s affordance is related to the social scenarios in which it is used and to people's expectations of a specific social role. The research questions (RQs) are as follows. 
\begin{itemize}
    \item \textbf{RQ1}. What preferences do people have for a social robot's affordance in general?  
    \item \textbf{RQ2}. Do such preferences vary when people know what the robot may be used for? 
    \item \textbf{RQ3}. How do people's expected social perceptions of a social robot affect their preference for a social robot's affordance?
    \begin{itemize}
        \item RQ3a. What levels of warmth and competence do people expect from a social robot in different roles? 
        \item RQ3b. How do the expected warmth and competence level affect people's preferences for robots' looks?
        \item RQ3c. How do the expected warmth and competence level affect people's preferences for robots' voices?
    \end{itemize}
\end{itemize}

\section{Research Methods}
To answer the above questions, an experiment is conducted by means of a questionnaire. It asks about participants’ demographic information and experience with speech-enabled technologies. It also asks for participants’ expectations for a speaking social robot’s look and voice under different circumstances (1) without any information about what the robot is used for; (2) with the information provided about the robot’s role in six social scenarios, which includes three public and three private scenarios. The robot's role in each scenario is also provided. They are a receptionist robot at a museum, a leisure robot at home, a private health-care robot, a private language teacher robot, a waiter robot at a restaurant and a finance adviser robot at a bank. The full description of the scenarios is attached in the appendix. 

A 5-point Likert scale is adopted, on a scale of 1-5, with `1’ as `the least’ and `5’ as `the most’. Participants are asked to indicate how much they agree with the statements (e.g., `A robot's look should be as human-like as possible.') when they do not know what the robots are used for. Later, when provided with scenario information, participants are asked to indicate their preferred level (e.g., `The robot needs to be competent.', `The robot needs to have a human-like look.').

Given that there are four variables to measure in six scenarios, and each participant goes through each condition (within groups), the statistical software G*Power is used to conduct a prior test to determine the required sample size. By selecting the ANOVA test (repeated measures), with the desired large effect size of 0.5, alpha of 0.05, power of 0.8, a moderate correlation of 0.5, and the type of effect size measure being `Partial Eta Squared', a total sample size of at least 42 participants is required. 

The study is a part of the empirical human-robot interaction experiment, which has received ethical approval from the Department of Computer Science at the University of Sheffield.

\section{Current Results}
Up to the revision date (28 February 2023), 68 participants completed the survey. The experiment is expected to end in early March. All survey results so far are valid to be processed. According to the survey results, 35.3\% of the participants are aged between 18-24 years old, 39.7\% of them are 25-34 years old, 13.2\% of them are 35-44 years old and 11.8\% of them are above 45 years old. The gender distribution is 57.4\% female, 35.3\% male and 7.4\% non-binary. Over 60\% of the participants are British. Nearly 20\% of participants report that their English accents `occasionally cause problems in daily communication'. Most participants claim they do not know much about speech science (72\%) or conversational artificial intelligence (55\%). 

For \textbf{RQ1}, when participants do not know what the social robot would be used for, they generally prefer it to have a less human-like look (mean 1.9), a more human-like voice (mean 3.1) and less human-like behaviour (mean 2.3). After performing a one-way analysis of variance (ANOVA), it shows these preferences have a statistically significant difference (p-value = 0.000). As shown in Figure 1, the violin plot shape shows the probability density of each set, and the box plot shows the range and means of the agreement level.

\begin{figure}[h]
  \centering
  \includegraphics[width=\linewidth]{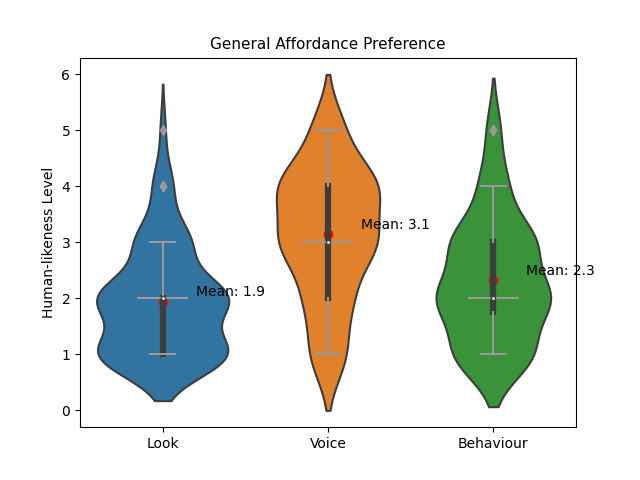}
  \caption{The graph shows people's preferences for a generic social robot's affordance, with the highest preference for a human-like voice, followed by human-like behaviours and a human-like look.}
  \Description{People's preference for a social robot's look, voice and behaviour is shown in a combination of box and violin plots.}
\end{figure}

For \textbf{RQ2}, it is found that, in comparison with general preferences, participants' preference for a social robot's affordance differs among the six social roles provided. As shown in Table 1, participants still prefer a more human-like voice than a human-like look, no matter which situation a robot is used in. Also, participants' preference for a human-like look increases slightly when they know what social robots are used for. Still, people hold different opinions about how human-like it needs to be according to the robot’s roles in given social scenarios. After performing a one-way analysis of variance (ANOVA), it shows (1) there is no significant difference between the sets of human-like looks (p=0.354, >0.05); (2) there is a significant difference between the sets of human-like voices (p-value=0.000). 

\begin{table}[h]
  \caption{Situational Affordance Preference}
  \label{tab:situational preferences}
    \begin{tabular}{ c|cc|cc }
    \toprule
     Robot roles & \multicolumn{2}{c}{Human-like Look} & \multicolumn{2}{c}{Human-like Voice}\\
      & mean & std. & mean & std.\\
     \hline 
     general use & 1.9 & 1.0 & 3.1 & 1.2\\
     \hline 
     receptionist & 2.1 & 1.1 & 3.1 & 1.2 \\
     leisure companion & 2.2 & 1.1 & 3.5 & 1.4\\
     home-carer & 2.2 & 1.2 & 2.9 & 1.3 \\
     language teacher & 2.1 & 1.1 & 4.3 & 1.1\\
     waiter & 2.3 & 1.1 & 3.0 & 1.2\\
     financial adviser & 2.3 & 1.2 & 3.1 & 1.2\\
    \bottomrule
    \end{tabular}
\end{table}

To answer \textbf{RQ3}, further analysis is conducted to investigate what roles social perception (warmth and competence) plays in people's preferences of social robots' affordance. For \textbf{RQ3a}, as shown in Table 2, participants expect a medium-high warmth of all social roles. Compared to that, participants hold higher expectations of a social robot's competence. As for \textbf{RQ3b} and \textbf{RQ3c}, which is about potential correlative relations between expected warmth, competence and human-like look and voice, a correlative matrix is produced (Figure 2). It shows that the absolute correlation coefficients between social dimensions (warmth and competence) and a robot's affordance (look and voice) are mostly less than 0.5. The degree of a human-like look is negatively connected with competence and positively associated with warmth at a small level. The degree of human-like voice is positively related to competence at a small level and negatively correlated with warmth at a moderate level.  
\begin{table}[t]
  \caption{Expected Social Perception of Situational Robot Roles}
  \label{tab:situational preferences}
    \begin{tabular}{ |c|c|c|c|c| }
    \hline
     robot roles & \multicolumn{2}{c}{warmth} & \multicolumn{2}{c}{competence}\\
       & mean & std. & mean & std.\\
     \hline
     receptionist & 3.8 & 0.9 & 4.7 & 0.7 \\
     leisure companion & 4.3 & 1.0 & 4.4 & 0.8\\
     home-carer & 3.7 & 1.3 & 4.7 & 0.8 \\
     language teacher & 3.7 & 1.2 & 4.8 & 0.7\\
     waiter & 3.8 & 1.2 & 4.6 & 0.7\\
     financial adviser & 3.5 & 1.3 & 4.8 & 0.7\\
     \hline
    \end{tabular}
\end{table}

\begin{figure}[h]
  \centering
  \includegraphics[width=\linewidth]{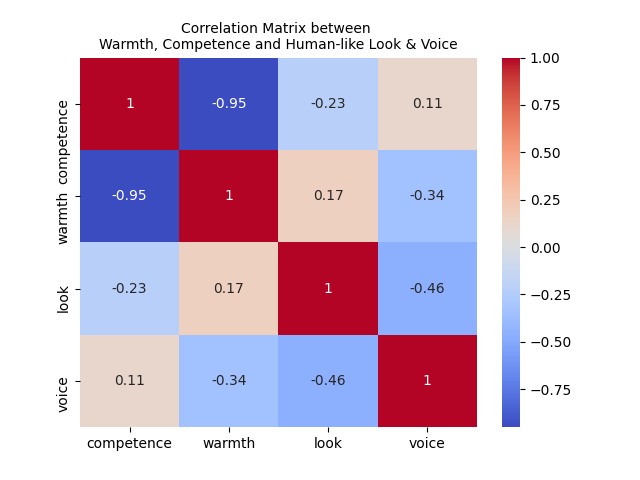}
  \caption{The heatmap shows relations between people's expectation of a robot and their preference for its affordance.}
  \Description{Most of the heatmap is the cold colours or pale yellow colour.}
\end{figure}

\section{Discussion and Conclusion}
The human-likeness of social robots is a controversial topic. The current experiment shows that human-like affordances are not necessarily favoured. People hold a medium-low preference for the human-like affordance in a social robot, with a higher preference for a human-like voice. These general preferences stay the same in situational conditions. The results of this study are based on people's subjective preferences, not their perception of real social robots. Why do people not hold a strongly positive view of human-like affordance? This study does not provide an answer to that. It could be caused by people's concerns about the potential `Uncanny Valley Effect', which states the more human-like an object is, the more affinity it gains until a point where the eerie sensation occurs \cite{mori2012uncanny}. It could be rooted in people's ethical concern about mixing boundaries between humans and artificial agents or their preference for a transparent and helpful affordance to reduce potential uncertainties when they interact with the social robot. Further studies are needed to develop a better understanding of the causes. 

Another aspect to discuss is that a social robot's look and voice are not preferred to be 100\% aligned. In other words, a less human-like look and a more human-like voice may work. The questions are, what the safe boundary of such difference is and how we can measure it. If the difference between a social robot's voice and look is too big, it may generate mismatched perceptual cues, which can trigger an uncanny feeling \cite{moore2012bayesian,meah2014uncannymisalignedcues,katsyri2015review,macdorman2016reducing}. Thus, whilst it is sensible to consider individual components of a social robot's affordance, its overall affordance should also be considered.

Furthermore, the degree of human-likeness differs dependent on the scenarios. Take a robot's voice for example. Whilst a moderate human-like voice (3.0 +/- 0.1) is acceptable under most circumstances, people's opinions on a robot's voice could differ a lot in some scenarios: an almost human-like voice is preferred for a language teacher role. This may be related to the professional requirements of the role. Thus, it is important to gain further understandings of people's many different needs in scenarios.

As for the role of social perception, the results show some correlation between people's expectations of how warm and competent a social robot should be and how it should look and sound. Such correlations are the opposite. For example, if people would like a more competent social robot in a given situation, they may like a slightly more human-like voice and a slightly less human-like look. Given that the warmth and competence levels are measured by one stated preference only and people may have different understandings of warmth and competence, it is worth investigating further to enhance the reliability of the result. Also, it will be interesting when the project proceeds to see whether the expected social perceptions match real-life social perceptions.

Furthermore, one essential element in HRI is a social robot's behaviour. From the current survey result, people do not expect a highly human-like behaviour pattern in a social robot. There is one question to follow up on. That is, for the robot used in the same social situation and having the same behaviour patterns, how do affordance designs affect people’s perception of the role and their interaction experience? To answer this question, an extended in-lab session of the above experiment is conducted. In the session, the same group of participants is invited to have a face-to-face interaction with a speaking social robot in the human-robot-interaction lab. The social robot `Furhat' is used to interact with participants. It plays three roles in three interactions: a quiz host, a joke maker and a scientist (within the subject). Each role comes with three affordance settings: robot-like, human-like (adult) and human-like (child) (between subjects). Roles' order and affordance are randomised to eliminate the sequence bias. Participants need to complete rating tasks before and after their interactions and elaborate on their thoughts during post-interaction interviews. Up to the submission date, 
45 participants completed the in-lab session. It is too early to report the effect of different looks and voices on the same robot roles. However, post-interaction interviews show that individual users have different perceptions of the same robot’s language behaviour. Taking the word `sweet’ (as a response when users agree to have another joke) as an example, some participants find it too childish and not sincere, whilst others find it light-hearted. Participants also pointed out that the sociolinguistic cues contained in the voice could affect their willingness to interact with certain robotic characters.  

Yet to be completed, the empirical data from the experiment presented in this paper shows that personalised affordance design is critical in human-robot interaction and should be considered part of a comprehensive approach to optimising human-robot interaction. It also shows that personalised affordance design takes place within a multi-dimensional space. It requires some design principles that take social factors into consideration. This report only covers one aspect of the experiment conducted so far. It is anticipated to be a starting point for further discussions, such as which features make a robot's look and voice more likeable. Added to this is how participants' previous experience and expectations affect their preference and perception of a social robot's affordance.

\begin{acks}
This work was supported by the Centre for Doctoral Training in Speech and Language Technologies (SLT) and their Applications, funded by UK Research and Innovation [grant number EP/S023062/1].
\end{acks}

\bibliographystyle{ACM-Reference-Format}
\balance
\bibliography{mybib}

\appendix 
\section{Six Social Scenarios}
\begin{enumerate}
            \item You are visiting a museum. There is a receptionist robot that can answer your questions about items in the exhibition and the museum's history. 
            \item You are living alone at home. There is a leisure robot that can read your stories, sing songs and play video clips with your photos. 
            \item You are ill. There is a service robot that can bring you medicine and do housework for you. 
            \item You are learning a new language. There is a multi-lingual robot that can demonstrate how to speak a word and show you what goes wrong with your pronunciation.
            \item You are visiting a newly open restaurant, and there is a waiter robot. It can take your order and bring you food. 
            \item You are visiting a bank to sort out your mortgage. There is a banker robot which can evaluate your situation and give you advice. 
\end{enumerate}

\end{document}